\documentclass[runningheads]{llncs}
\usepackage[T1]{fontenc}
\usepackage{graphicx,verbatim}
\usepackage{amsmath,amssymb}
\usepackage{adjustbox}
\usepackage{tikz}

\usetikzlibrary{arrows.meta, positioning, shapes.geometric, fit, backgrounds, calc}

\definecolor{navy}{RGB}{31,58,94}
\definecolor{steel}{RGB}{70,114,158}
\definecolor{skyfill}{RGB}{228,237,247}
\definecolor{tealline}{RGB}{34,120,120}
\definecolor{tealfill}{RGB}{225,243,241}
\definecolor{amberline}{RGB}{176,108,20}
\definecolor{amberfill}{RGB}{251,238,216}
\definecolor{grayline}{RGB}{95,95,95}
\definecolor{grayfill}{RGB}{240,240,240}
\definecolor{oodline}{RGB}{158,44,44}
\definecolor{oodfill}{RGB}{250,227,227}

\begin{document}

\title{CalTwin: Towards Calibrated, Shift-Robust Medical World Models via Fisher-Information Regularisation}
\titlerunning{Calibrated, Shift-Robust Medical World Models}
\author{Behraj Khan\inst{1} \and Shabir Ahmad\inst{2} \and Syed Ahmad Chan Bukhari\inst{3} \and Tahir Qasim Syed\inst{1}}
\authorrunning{B. Khan et al.}
\institute{Institute of Business Administration Karachi, Karachi, Pakistan\\
\email{behrajkhan@iba.edu.pk, tqsyed@iba.edu.pk} \and
Gachon University, Korea\\ 
\email{shabir@ieee.org}\and
St. John's University, New York, USA\\
\email{bukharis@stjohns.edu }}

\maketitle
%
\begin{abstract}
Medical world models aim to learn a latent state of patient or organ
physiology and a transition function that forecasts how that state evolves
under interventions, supporting downstream tasks from imaging-based diagnosis
to digital-twin treatment planning. Two failure modes threaten the reliability
of such models in clinical deployment: (i)~\emph{covariate shift}, because
training data are fragmented across hospitals, scanners, and time, so the
feature distribution seen by the latent-dynamics predictor differs across
fragments and from the distribution at deployment; and (ii)~\emph{confidence
misalignment}, because multi-step forecasts are often overconfident exactly
where clinical risk is highest. We argue that both problems admit a unified
treatment via a single lightweight regularisation objective, \textbf{CalTwin},
which combines a Fisher-Information-based shift penalty adapted from our prior
work on fragmented covariate-shift remediation~\cite{khan2025mitigating,khan2025causal} with a
Confidence Misalignment Penalty adapted from our prior work on calibrated
vision-language classification~\cite{khan2025confidence}, applied here to a
GRU-based medical world model's latent transition predictor. We derive the
combined objective, establish which proof steps transfer from the
classification setting without modification and which require adaptation, and
evaluate it on the PhysioNet 2019 Sepsis Challenge, treating the two hospital
systems as sequential training fragments and the unseen system as an
out-of-distribution test. CalTwin reduces OOD next-step latent-state MSE
by 9.1\% relative to the no-penalty baseline (FIM penalty alone accounts
for 7.0\%); the ECE reduction from the Confidence Misalignment Penalty is
real but small (0.7\% for CalTwin, 1.3\% for CMP alone). We are explicit
about what the present experiments do and do not establish, and identify the
steps needed to validate CalTwin on the imaging modalities central to this
workshop.

\end{abstract}

\section{Introduction}

The emerging paradigm of medical world models learns a compact latent
representation of a patient's or organ's physiological state together with a
transition function that predicts how that state evolves, conditioned on
time, interventions, or imaging acquisitions. A recent survey formalises the
target distribution as $p(s_{t+1} \mid s_t, a_t)$ and surveys implementations
spanning longitudinal MRI simulation, radiograph projection dynamics, EHR
trajectory generation, cardiac guidance, and surgical video synthesis,
placing methods on a four-level capability ladder from temporal prediction
(L1) through action-conditioned simulation (L2) to counterfactual rollouts
(L3) and closed-loop planning (L4)~\cite{qazi2025beyond}, noting
that L3--L4 capabilities  the ones of highest clinical value  remain rare.
Medical digital twins  five-component systems comprising the patient, a
data connection, a \emph{patient-in-silico}, a clinical interface, and a
twin-synchronisation mechanism~\cite{sadee2025medical}  are the natural L3--L4
embodiment of this paradigm, and both~\cite{sadee2025medical} and a healthcare
digital-twin review~\cite{rudsari2025digital} identify federated learning and
trustworthy, multimodal data integration as the open priorities standing
between prototypes and clinical deployment.

Two properties of the medical setting make trustworthy deployment of such
systems difficult to achieve in practice, and we argue that they are best
treated as a unified technical problem rather than separately.

\textbf{Covariate shift across fragmented, non-colocated training data.}  A
medical world model's transition predictor is rarely trained on a single,
centrally held dataset: patient trajectories are fragmented across hospitals,
scanner vendors, protocols, and time, and privacy regulation (HIPAA, GDPR)
pushes training toward federated or batch-sequential regimes. Each fragment
then has its own empirical distribution $P_k(s_t)$, exposing the latent
dynamics model to a sequence of covariate distributions that differ from one
another and from the deployment distribution. This is not hypothetical: for
sepsis and ICU mortality prediction on the eICU and MIMIC-III benchmarks,
federated models trained on one set of hospitals consistently underperform on
held-out hospitals due to demographic and clinical-practice covariate
shifts~\cite{zhu2025fedweight}. For a world model, this fragment-level shift acts
on every step of a predicted trajectory, and the resulting per-step error
compounds across the forecast horizon.

The classical response, importance weighting by an estimated density ratio
$P_{\mathrm{dep}}(s)/P_{\mathrm{train}}(s)$~\cite{shimodaira2000improving}, is
inapplicable here: the deployment distribution is unknown at training time,
and in a federated/batch-sequential regime both the training and reference
distributions change across fragments, so no single density ratio corrects
for all fragment-level shift at once. A parameter-space regulariser that
penalises drift from an information-geometry-based prior  the Fisher
Information Matrix (FIM) accumulated across fragments  avoids both problems:
it needs only the model's own gradients on the current fragment and updates
incrementally as fragments arrive. This is the approach formalised in our
prior work on Fragmentation-Induced Covariate Shift Remediation
(FIcsR)~\cite{khan2025mitigating} and its preliminary version~\cite{khan2025causal}, which we
adapt here to the transition predictor of a medical world model.

\textbf{Confidence misalignment in multi-step clinical forecasts.}
Even with fragment-level shift controlled, a world model deployed
autoregressively  conditioning each step on its own previous output  faces
a second, compounding failure mode. Training uses teacher forcing (true,
clinician-acquired states as input); at deployment the model conditions on
its own prior predictions, and this mismatch accumulates across the forecast
horizon. The predictive confidence reported at step $t$ reflects uncertainty
given \emph{true} previous states, not the model's own increasingly erroneous
self-generated ones. Neural classifiers are already overconfident
in-distribution~\cite{guo2017calibration}, and post-hoc fixes such as temperature
scaling are known to \emph{worsen} calibration under dataset
shift~\cite{snoek2019can}  ruling out a post-hoc solution, since the shift
here is endogenous to the model's own prediction process. An overconfident
wrong forecast (e.g., of a treatment response in a cardiac digital twin) is
more dangerous than an honestly uncertain one, because the clinician has no
signal from the model's own confidence that the output should be discounted.

\textbf{Our position.} Both problems admit unified treatment via one family
of Fisher-Information-based regularisers whose components are already
separately validated for classification on non-medical data:
FIcsR~\cite{khan2025mitigating,khan2025causal} penalises KL divergence between a fragment's
parameter distribution and an accumulated FIM-based global prior, improving
generalisation under fragment-level shift; CalShift~\cite{khan2025confidence}
augments this with CMP, redistributing log-likelihood away from
overconfident incorrect predictions, jointly improving accuracy and ECE for
CLIP-based classifiers under shift, with both penalties provably combining
into one objective. We adapt both, together, to a medical world model's
latent transition predictor  FIM for cross-fragment shift in the dynamics,
CMP for calibration across an autoregressive trajectory  specifying which
derivation steps transfer directly, which require modification, and what
validation remains (Sec.~4), on real multi-hospital ICU time series treated
as a proxy benchmark rather than a substitute for imaging validation.

\section{Related Work}

\textbf{Medical digital twins and world models.} Digital twins originate in engineering~\cite{sadee2025medical}; their translation to medicine is fast-growing but lacks a settled architectural consensus~\cite{sadee2025medical}. Healthcare digital-twin surveys identify multimodal data integration, federated learning, and trustworthy/explainable modeling as open priorities~\cite{rudsari2025digital}, and generative-AI human-digital-twin reviews note that high-fidelity modeling must typically be built from scarce, biased, multi-source data~\cite{chen2024generative}. The world-model paradigm  learning a latent state and transition predictor jointly  has recently been proposed for clinical prediction, counterfactual reasoning, and planning across imaging, EHR, and surgical-robotics use cases~\cite{qazi2025beyond}; we treat its latent dynamics predictor as the object that must be made shift-robust and calibrated.

\textbf{Covariate shift correction.} Classical approaches reweight training examples by an estimated train/test density ratio~\cite{khan2025mitigating}; Moreno-Torres et al.\ showed $k$-fold cross-validation itself induces such shift~\cite{khan2025mitigating}. FIcsR and its preliminary version C$^3$ instead approximate the relative entropy between a fragment's distribution and a global reference via the empirical Fisher Information Matrix, tractable for over-parametrised networks and accumulable incrementally across batches/folds without joint access to both distributions' samples~\cite{khan2025mitigating,khan2025causal}  the mechanism we adapt to cross-site, cross-time medical world model fragmentation.

\textbf{Confidence calibration under distribution shift.} Calibration is known to degrade under distribution shift and low-data fine-tuning~\cite{khan2025confidence}. CalShift combines the FIM penalty above with a CMP term that moves probability mass away from incorrect classes in proportion to how much they exceed the true class's probability, provably bounded in $[0,1]$ and vanishing for a well-calibrated model~\cite{khan2025confidence}. This combined objective has not, to our knowledge, previously been proposed for a medical world model's multi-step latent dynamics predictor  the gap this paper addresses.

\textbf{Multi-hospital shift in clinical time series.} Independent of the world-model literature, federated-EHR work has directly quantified the shift our Sec.~4 experiment targets: models trained on eICU hospitals show degraded mortality/sepsis prediction on the disjoint MIMIC-III hospital system, attributable to demographic and practice differences rather than task difficulty~\cite{zhu2025fedweight}. This motivates our use of the (similarly multi-hospital, real-not-synthetic) PhysioNet 2019 Sepsis Challenge~\cite{reyna2019early} as a covariate-shift testbed in Sec.~4.
\section{Proposed Method}

\subsection{Setting: the latent transition predictor of a medical world model}

We adopt the world model formulation surveyed in~\cite{qazi2025beyond}:
an encoder $\mathcal{E}$ maps multimodal clinical inputs at time $t$
(imaging volumes, physiological signals, or EHR features) to a latent state
$s_t = \mathcal{E}(x_t)$, and a parametric transition predictor $f_\theta$
models $p_\theta(s_{t+1} \mid s_t, a_t)$, with $a_t$ an optional
intervention/acquisition variable. The only property of $f_\theta$ we
require is that $\log p_\theta(s_{t+1}\mid s_t, a_t)$ be differentiable in
$\theta$.

Training data come from $K$ fragments $\{B_1,\ldots,B_K\}$, each a hospital
site, scanner vendor, protocol, or time window, with empirical distribution
$\hat P_k(s_t)$; $P_{\mathrm{dep}}(s_t)$ denotes the unknown deployment
distribution. The objective must produce $\theta$ that (i) generalises across
fragments, (ii) generalises to $P_{\mathrm{dep}}$, and (iii) gives
well-calibrated confidence at each step of an autoregressive rollout.

\begin{figure*}[t]
\centering
\sffamily
\resizebox{\textwidth}{!}{
\begin{tikzpicture}[
  font=\sffamily\footnotesize,
  node distance=10mm and 10mm,
  >={Stealth[length=2mm,width=1.6mm]},
  every node/.style={align=center},
  dbicon/.style={cylinder, draw=steel, shape border rotate=90, aspect=0.22,
                 minimum height=13mm, minimum width=11mm, fill=skyfill, thick},
  oodicon/.style={cylinder, draw=oodline, dashed, shape border rotate=90, aspect=0.22,
                 minimum height=13mm, minimum width=11mm, fill=oodfill, thick},
  encicon/.style={trapezium, draw=steel, thick, fill=skyfill,
                 trapezium left angle=68, trapezium right angle=112,
                 minimum width=17mm, minimum height=13mm},
  gruicon/.style={draw=steel, thick, rounded corners=2pt, fill=skyfill,
                 minimum width=21mm, minimum height=12mm},
  smallblock/.style={draw=grayline, thick, fill=grayfill, rounded corners=2pt,
                minimum height=9mm, text width=22mm, inner sep=1.5mm},
  regblock/.style={draw=tealline, thick, fill=tealfill, rounded corners=2pt,
                minimum height=15mm, text width=25mm, inner sep=1.8mm},
  objblock/.style={draw=amberline, very thick, fill=amberfill, rounded corners=3pt,
                minimum height=15mm, text width=54mm, inner sep=2.2mm},
  evalblock/.style={draw=steel, thick, fill=white, rounded corners=2pt,
                text width=32mm, inner sep=2.5mm},
  panel/.style={draw=grayline!70, rounded corners=4pt, inner sep=4mm, fill=none},
  paneltitle/.style={font=\sffamily\bfseries\scriptsize, text=navy},
  arr/.style={-{Stealth}, thick, draw=grayline!85!black},
  darr/.style={-{Stealth}, thick, dashed, draw=amberline},
]

\node[dbicon] (f1) at (0,0) {\scriptsize $B_1$};
\node[dbicon] (f2) [right=9mm of f1] {\scriptsize $B_2$};
\node[dbicon] (f3) [right=9mm of f2] {\scriptsize $B_3$};
\draw[arr] (f1) -- (f2);
\draw[arr] (f2) -- (f3);

\node[oodicon] (ood) [below=13mm of f2] {\scriptsize Hosp.\ B};
\draw[oodline, thick] ($(ood.north)+(0,3.3mm)$) circle (1.7mm);
\draw[oodline, thick] ($(ood.north)+(-1.1mm,2.2mm)$) -- ($(ood.north)+(1.1mm,4.4mm)$);
\node[oodline, font=\scriptsize\itshape, below=0.5mm of ood] {held out};

\node[encicon] (enc) [right=20mm of f3] {\scriptsize Encoder\\[-0.3mm]$\mathcal{E}$};
\node[gruicon] (gru) [right=22mm of enc] {\scriptsize GRU transition\\[-0.3mm]$p_\theta(s_{t+1}\mid s_t)$};
\draw[-{Stealth}, thick, steel] ($(gru.north)+(-2mm,2.2mm)$) arc (0:320:2mm);

\draw[arr] (f3.east) -- (enc.west);
\draw[arr] (enc) -- (gru) node[midway, above, font=\scriptsize] {$s_t$};

\node[evalblock] (ev) [right=26mm of gru] {\textbf{Evaluation}\\[1.5mm]
  {\scriptsize ID test: held-out Hosp.\,A}\\[1mm]
  {\scriptsize OOD test: Hospital B}\\[1mm]
  {\scriptsize Metrics: MSE, ECE, AUROC}};

\draw[arr] (gru.east) -- (ev.west);

\draw[arr, oodline, dashed] (ood.east) -| (ev.south);

\node[smallblock] (aux) [below=15mm of enc, xshift=18mm] {Auxiliary\\risk head};
\draw[grayline, thick, fill=white] ($(aux.north east)+(-1.2mm,-1.2mm)$) circle (1.5mm);
\draw[grayline, thick] ($(aux.north east)+(-1.2mm,-1.2mm)$) circle (0.6mm);
\node[smallblock, text width=18mm] (yhat) [below=8mm of aux] {Predicted\\risk $\hat y_t$};

\draw[arr] (enc.south) -- ++(0,-6mm) -| (aux.north);
\draw[arr] (aux) -- (yhat);

\node[regblock, minimum height=22mm] (fim) [below=10mm of yhat, xshift=-18mm] {\scriptsize Fisher-information\\ $\lambda_1\mathcal{L}_{\mathrm{FIM}}$  \\[1.2mm]
  \begin{tikzpicture}[baseline]
    \foreach \i in {0,1,2}{
      \foreach \j in {0,1,2}{
        \pgfmathsetmacro{\shadeval}{25+9*\i+5*\j}
        \draw[fill=tealline!\shadeval, draw=tealline!65]
          (\i*2.1mm,\j*2.1mm) rectangle ++(1.9mm,1.9mm);
      }
    }
    \draw[thick] (-1.1mm,-1.1mm) -- (-1.1mm,7.5mm) -- (-0.2mm,7.5mm);
    \draw[thick] (-1.1mm,-1.1mm) -- (-0.2mm,-1.1mm);
    \draw[thick] (7.5mm,-1.1mm) -- (7.5mm,7.5mm) -- (6.6mm,7.5mm);
    \draw[thick] (7.5mm,-1.1mm) -- (6.6mm,-1.1mm);
  \end{tikzpicture}\\[0.3mm]
  {\scriptsize $\hat I^{(k)}(\theta)$}
};
\node[regblock] (cmp) [right=12mm of fim] {Confidence Missalignement Penalty\\$\lambda_2\,\mathrm{CMP}$};
\node[objblock] (obj) [right=12mm of cmp] {\textbf{CalTwin training objective}\\[1mm]
  };

\draw[arr] (f2.south) -- ++(0,-10mm) -| (fim.north);
\draw[arr] (yhat.south) -- ++(0,-4mm) -| (cmp.north);
\draw[arr] (fim.east) -- (cmp.west);
\draw[arr] (cmp.east) -- (obj.west);
\draw[arr] (gru.south) -- ++(0,-6mm) -| node[midway, right, font=\scriptsize] {NLL} (obj.north);

\coordinate (loopBot)  at ($(obj.south)+(0,-8mm)$);
\coordinate (loopLeft) at ($(f1.west)+(-12mm,0)$);
\coordinate (topY)     at ($(f1.north)+(0,9mm)$);
\draw[darr]
  (obj.south) -- (obj.south |- loopBot) --
  (loopLeft |- loopBot) --
  (loopLeft |- topY) --
  (enc.north |- topY) -- (enc.north);
\node[font=\scriptsize, text=amberline, fill=white, inner sep=0.5mm]
  at (loopLeft |- topY) [anchor=south, xshift=14mm, yshift=-1mm] {update $\theta$ (Eq.~6)};

\node[paneltitle] at ($(f1.north west)+(0,4mm)$) [anchor=south west] {Fragmented multi-site data};
\node[paneltitle] at ($(enc.north west)+(0,4mm)$) [anchor=south west] {Medical world model};
\node[paneltitle] at ($(ev.north west)+(0,4mm)$) [anchor=south west] {Evaluation};
\node[paneltitle] at ($(fim.south west)+(0,-4mm)$) [anchor=north west] {Calibrated regularization};

\begin{scope}[on background layer]
  \node[panel, fit=(f1)(f3)(ood), fill=skyfill!30] {};
  \node[panel, fit=(enc)(gru), fill=steel!6] {};
  \node[panel, fit=(ev), fill=skyfill!30] {};
  \node[panel, fit=(fim)(cmp)(obj), fill=tealfill!35] {};
\end{scope}

\begin{scope}[shift={($(fim.south west)+(0,-16mm)$)}, font=\scriptsize]
  \draw[arr] (0,0) -- (7mm,0); \node[anchor=west] at (8mm,0) {data / feature flow};
  \draw[darr] (32mm,0) -- (39mm,0); \node[anchor=west] at (40mm,0) {parameter update ($\theta$)};
  \draw[oodline, dashed, thick] (68mm,0) -- (75mm,0); \node[anchor=west] at (76mm,0) {held-out OOD path};
\end{scope}

\end{tikzpicture}
}
    \caption{The CalTwin training pipeline. Sequential hospital fragments
    $B_1,B_2,B_3$ train the shared encoder $\mathcal{E}$ and GRU transition
    predictor (Sec.~3.1); a held-out hospital is used only for OOD evaluation.
    $s_t$ also feeds an auxiliary risk head penalised by CMP (Sec.~3.3); a
    diagonal Fisher-information accumulator anchors encoder/transition
    parameters against prior fragments (Sec.~3.2). Both penalties and the
    transition log-likelihood combine into the CalTwin objective
    (Eq.~\ref{eq:lcaltwin}), whose gradient updates $\theta$ (dashed loop).}
    \label{fig:method}
\end{figure*}
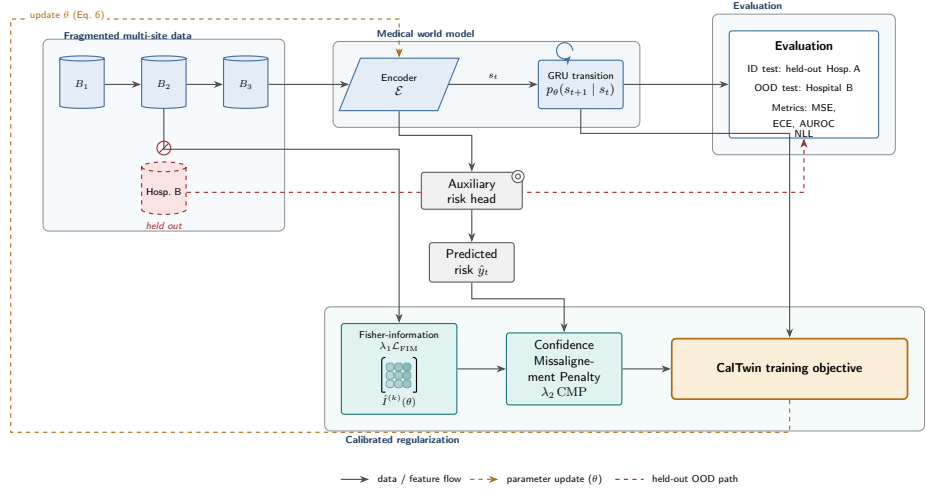

\subsection{Fisher-Information regularisation for cross-fragment covariate
shift}

Following~\cite{khan2025mitigating,khan2025causal}, we approximate the KL divergence between
the parameter distribution $P(\theta \mid B_k)$ induced by training on
fragment $B_k$ and a global reference $Q(\hat\theta)$ via a second-order
Taylor expansion of the log-likelihood, which yields the Fisher Information
Matrix (FIM) as the leading term~\cite{ref_lecun2002}:
\begin{equation}
D_{\mathrm{KL}}\!\left(P(\theta\mid B_k)\,\Vert\,
Q(\hat\theta)\right)
\;\approx\;
\tfrac{1}{2}(\hat\theta - \theta)^{\!\top}
I(\theta)\,(\hat\theta - \theta),
\qquad
I(\theta) = -\,\mathbb{E}\!\left[
\frac{\partial^2 \log P(X\mid\theta)}{\partial\theta\,\partial\theta^{\!\top}}
\right].
\label{eq:fim}
\end{equation}
This is the same approximation used in Elastic Weight Consolidation
(EWC)~\cite{kirkpatrick2017overcoming} and in FIcsR~\cite{khan2025mitigating} for classification. For an
over-parametrised world model the full FIM is intractable, so we use the
empirical Fisher:
\begin{equation}
I_e(\theta) = \frac{1}{N}\sum_{i=1}^{N}
g_i g_i^{\!\top},
\qquad
g_i = \frac{\partial \log p_\theta(s_{i,t+1}\mid s_{i,t}, a_{i,t})}
           {\partial \theta},
\label{eq:empfim}
\end{equation}
computed over the transition log-likelihood of the world model rather than a
classifier's categorical log-likelihood. This is the primary adaptation
step relative to~\cite{khan2025mitigating,khan2025causal}, which define $g_i$ as the gradient
of a softmax cross-entropy loss; here $g_i$ is instead the gradient of
$\log p_\theta(s_{t+1}\mid s_t, a_t)$, which may be Gaussian, a masked
autoregressive flow, or a diffusion denoising objective. The positive
semi-definiteness of $I_e(\theta)$ and the validity of the KL approximation
(Eq.~\ref{eq:fim}) are form-independent  both follow from $I_e$'s
outer-product structure and log-likelihood smoothness, not from the specific
form of $p_\theta$  so the proofs in~\cite{khan2025mitigating} carry over to the
transition-predictor setting without modification; we flag this explicitly
to separate established results from extensions.

A global accumulated FIM is maintained across fragments in an
exponential-moving-average fashion:
\begin{equation}
\hat{I}^{(k)} = \alpha\,\hat{I}^{(k-1)} + (1-\alpha)\,I_e^{(k)}(\theta),
\label{eq:accumfim}
\end{equation}
where $\alpha \in (0,1)$ is a decay parameter that discounts older fragments,
and $I_e^{(k)}(\theta)$ is the empirical FIM computed on the current fragment
$B_k$. Each new fragment is then trained with the objective
\begin{equation}
\mathcal{L}_{\mathrm{shift}}(\theta)
= -\log p_\theta(s_{t+1}\mid s_t, a_t)
+ \lambda_1\,(\hat\theta - \theta)^{\!\top}
  \hat{I}^{(k-1)}\,(\hat\theta - \theta),
\label{eq:lshift}
\end{equation}
where $\hat\theta$ is the parameter estimate from the previous fragment. The
regulariser penalises movement, scaled by Fisher Information, away from
parameters that were optimal for all previous fragments  the
Cram\'er-Rao-bound anchoring mechanism of~\cite{khan2025mitigating} adapted to the
transition predictor. A diagonal approximation to $\hat{I}^{(k)}$, as used
in Sec.~4, is standard practice for tractability~\cite{kirkpatrick2017overcoming,khan2025mitigating}.

\subsection{Confidence Misalignment Penalty for autoregressive trajectories}

The second failure mode  predictive overconfidence compounding across
autoregressive steps  requires a training-time objective sensitive to the
shift from self-conditioning, not a post-hoc recalibration: temperature
scaling, while effective in-distribution~\cite{guo2017calibration}, worsens
calibration under exactly this kind of input-distribution
shift~\cite{snoek2019can}.

The Confidence Misalignment Penalty from~\cite{khan2025confidence} provides a
training-time alternative. For a discrete classification posterior over $C$
classes, CMP is defined as
\begin{equation}
\mathrm{CMP}(x, y)
= \frac{p_\theta(y \mid x)}
       {\displaystyle\sum_{\substack{y' \neq y \\
        p_\theta(y'\mid x) > p_\theta(y\mid x)}}
        p_\theta(y' \mid x)},
\label{eq:cmp}
\end{equation}
which is bounded in $[0,1]$, equals $1$ for a perfectly calibrated model, and
equals $0$ when the true class has zero predicted probability  penalising
probability mass assigned to classes that should not be preferred over the
true class, without a temperature parameter or held-out calibration set.

Adapting CMP to a world model requires addressing the fact that
$p_\theta(s_{t+1}\mid s_t, a_t)$ is continuous rather than a finite class set.
We assume the world model is jointly trained with one or more clinically
meaningful discrete auxiliary heads  e.g., a trajectory-risk classifier or a
binary treatment-response indicator  sharing the latent state $s_t$; such
heads are already standard in medical world models for downstream
supervision~\cite{qazi2025beyond}. For a head with posterior
$q_\theta(y\mid s_t)$ over $C$ outcomes, CMP is defined as in
Eq.~\ref{eq:cmp} with $p_\theta$ replaced by $q_\theta$, computed at each
trajectory step $t$. Because the head shares $\theta$ with the continuous
transition predictor via the encoder, calibration of the auxiliary head
propagates through the shared representation to the continuous
forecast~\cite{khan2025confidence}. This is a design choice, not a settled result;
a continuous-output generalisation via predictive variance is theoretically
attractive but left to future work. Sec.~4's auxiliary sepsis head, and its
discussion in Sec.~5, provide an initial empirical data point on how well
this discrete-head adaptation transfers under real cross-hospital shift.

\subsection{Combined objective}

The unified training objective for a medical world model's dynamics predictor,
trained across $K$ non-colocated fragments and required to produce calibrated
confidence on its auxiliary discrete outcome head, is
\begin{equation}
\mathcal{L}_{\mathrm{CalTwin}}(\theta)
= -\log p_\theta(s_{t+1}\mid s_t, a_t)
+ \lambda_1\,(\hat\theta - \theta)^{\!\top}\hat{I}^{(k-1)}(\hat\theta-\theta)
+ \lambda_2\,\mathrm{CMP}(s_t, y_t),
\label{eq:lcaltwin}
\end{equation}
directly mirroring $\mathcal{L}_{\mathrm{CalShift}} = \mathcal{L}_c +
\lambda_1 I(\theta) + \lambda_2\,\mathrm{CMP}$ in~\cite{khan2025confidence}, with
$\mathcal{L}_c$ replaced by the world model's transition log-likelihood. As
in~\cite{khan2025mitigating,khan2025confidence}, $\lambda_1,\lambda_2\ge0$ trade off
shift-robustness and calibration against base accuracy; our prior
classification work selected them via grid search on held-out
data~\cite{khan2025mitigating,khan2025confidence}, but, as made explicit in Sec.~4.2, the
experiments below use \emph{fixed} values rather than repeating that search,
so the reported numbers are a single point on the trade-off surface.

The two penalties are not independent: the FIM penalty bounds the
per-fragment shift in the representation $s_t=\mathcal{E}(x_t)$ that feeds
the auxiliary head, and reduced representation shift should ease CMP's
redistribution of probability mass. Conversely, CMP's overcommitted-mass
signal could in principle detect when $\lambda_1$ is insufficient, motivating
an adaptive re-weighting across fragments that we leave to future work.
Sec.~5 shows empirically that this independence assumption does not fully
hold: the two penalties interact measurably in the calibration domain.

\section{Experiments}

Sections~1--3 derive $\mathcal{L}_{\mathrm{CalTwin}}$ (Eq.~\ref{eq:lcaltwin})
in general; here we report a single-seed feasibility study on real
multi-site clinical time series, validated on two independent ICU datasets
(Secs.~4.1--4.4: PhysioNet 2019; Sec.~4.5: eICU-CRD Demo). This is a
proof-of-concept on tabular ICU surrogate tasks, not the ultrasound/cardiac-imaging
validation central to this workshop; we return to that gap in the Discussion.

\subsection{Dataset and fragmentation}

We use the PhysioNet 2019 Sepsis Challenge~\cite{reyna2019early} (hourly ICU
vital signs, two hospital systems A/B, binary sepsis label). From Hospital A
we take the first 3{,}600 subjects (patient-ID order, as an enrolment-order
proxy~\cite{khan2025mitigating}), split into $K{=}3$ chronological fragments of
1{,}200 (mirroring Sec.~3.2's batch-sequential arrival), holding out the last
15\% of each as an ID test split (1{,}020 train / 180 ID-test per fragment,
540 total). Hospital B is untouched during training: its first 1{,}200
subjects form the OOD test site (Sec.~1's cross-hospital shift, instantiated
directly~\cite{zhu2025fedweight}).

Each record uses seven vital-sign channels (heart rate, $\mathrm{SpO}_2$,
temperature, systolic/mean/diastolic BP, respiratory rate; $\mathrm{EtCO}_2$
dropped for near-total missingness). Missing values are forward/backward
filled, fully-missing channels imputed with a clinical reference value, and
trajectories truncated/padded to $T{=}24$ hourly steps and standardised
per-patient. Sepsis prevalence is 8.8\%/9.6\%/9.5\% for training fragments
0--2 and 5.6\% on the OOD site.

\subsection{Model and training details}

$f_\theta$ (Sec.~3.1) is: a linear encoder (LayerNorm, GELU, dropout $0.1$)
mapping 7 vitals to a $D{=}32$ latent $s_t$; a GRUCell ($32\rightarrow64$)
with two linear heads producing a diagonal Gaussian $p_\theta(s_{t+1}\mid
s_t)$; and an auxiliary MLP head ($32{\to}32{\to}1$) on the
trajectory-averaged latent state predicting sepsis onset, used for CMP
(Eq.~\ref{eq:cmp}). 24{,}385 trainable parameters. Adam ($\text{lr}=10^{-3}$,
weight decay $10^{-4}$), batch 32, gradient clipping at 1.0, 15 epochs/fragment,
trained sequentially over the 3 Hospital-A fragments (seed 42, single GPU).
The diagonal empirical FIM (Eq.~\ref{eq:empfim}) is recomputed after each
fragment (over 6 rollout steps/batch, for tractability) and EMA-accumulated
(Eq.~\ref{eq:accumfim}, $\alpha{=}0.9$). We use fixed $\lambda_1{=}0.5$,
$\lambda_2{=}0.3$ rather than grid search; tuning is left to future work.

Two details matter for Table~\ref{tab:results}. First, the auxiliary head is
trained \emph{only} through the CMP term  there is no separate
cross-entropy loss  so when $\lambda_2{=}0$ (Baseline, FIM-only) it receives
no gradient and stays at random initialisation; its AUROC of exactly
$0.5000$ reflects that design, not a discriminative failure (the informative
comparison is CMP-only vs.\ CalTwin; Sec.~5). Second, training was stable
across all three sequentially-arriving fragments for every method: CalTwin's
NLL decreased monotonically as fragments arrived ($-0.196\to-0.304\to-0.386$),
and for FIM-only/CalTwin the raw FIM penalty stayed $\le\!10^{-4}$ throughout
(it is exactly $0$ for Baseline/CMP-only by construction, since the FIM
accumulator is never populated when $\lambda_1{=}0$).

We compare four regimes (zeroing the corresponding penalty in
Eq.~\ref{eq:lcaltwin}): \textbf{Baseline} ($\lambda_1{=}\lambda_2{=}0$),
\textbf{FIM-only} ($\lambda_1{=}0.5,\lambda_2{=}0$), \textbf{CMP-only}
($\lambda_1{=}0,\lambda_2{=}0.3$), \textbf{CalTwin} ($\lambda_1{=}0.5,
\lambda_2{=}0.3$), reporting on both ID/OOD splits: mean next-step
latent-state MSE (23 teacher-forced rollout steps  i.e.\ conditional
accuracy/calibration under fragment/site shift, not closed-loop
autoregressive error); ECE (10 bins) and AUROC on the auxiliary sepsis head.

\subsection{Results}

\begin{table}[t]
\centering
\caption{Next-step latent-state MSE, calibration (ECE), and auxiliary-head
AUROC on the PhysioNet 2019 Sepsis Challenge, in-distribution (ID, pooled
held-out patients from all three Hospital-A fragments) and out-of-distribution
(OOD, Hospital B, unseen during training). Single seed. $\downarrow$: lower is
better; AUROC $\uparrow$: higher is better (0.5 = chance). Recall (Sec.~4.2)
that the auxiliary head is trained only when $\lambda_2>0$, so AUROC-ID/OOD
$=0.5000$ under Baseline/FIM-only reflects an \emph{untrained} head, not a
failed one.}
\label{tab:results}
\resizebox{\textwidth}{!}{
\begin{tabular}{lcccccc}
\hline
Method & MSE-ID $\downarrow$ & MSE-OOD $\downarrow$ & ECE-ID $\downarrow$ & ECE-OOD $\downarrow$ & AUROC-ID $\uparrow$ & AUROC-OOD $\uparrow$ \\
\hline
Baseline (no penalty)      & 0.2401 & 0.2614 & 0.4222 & 0.4442 & 0.5000 & 0.5000 \\
FIM only ($\mathcal{L}_{\mathrm{shift}}$) & 0.2232 & 0.2432 & 0.4222 & 0.4442 & 0.5000 & 0.5000 \\
CMP only                   & 0.2352 & 0.2587 & 0.4162 & 0.4384 & 0.5731 & 0.4984 \\
\textbf{CalTwin (FIM+CMP)} & \textbf{0.2228} & \textbf{0.2377} & 0.4189 & 0.4411 & 0.5740 & 0.4754 \\
\hline
\end{tabular}
}
\end{table}

\begin{figure}[!ht]
    \centering
    \includegraphics[width=\linewidth]{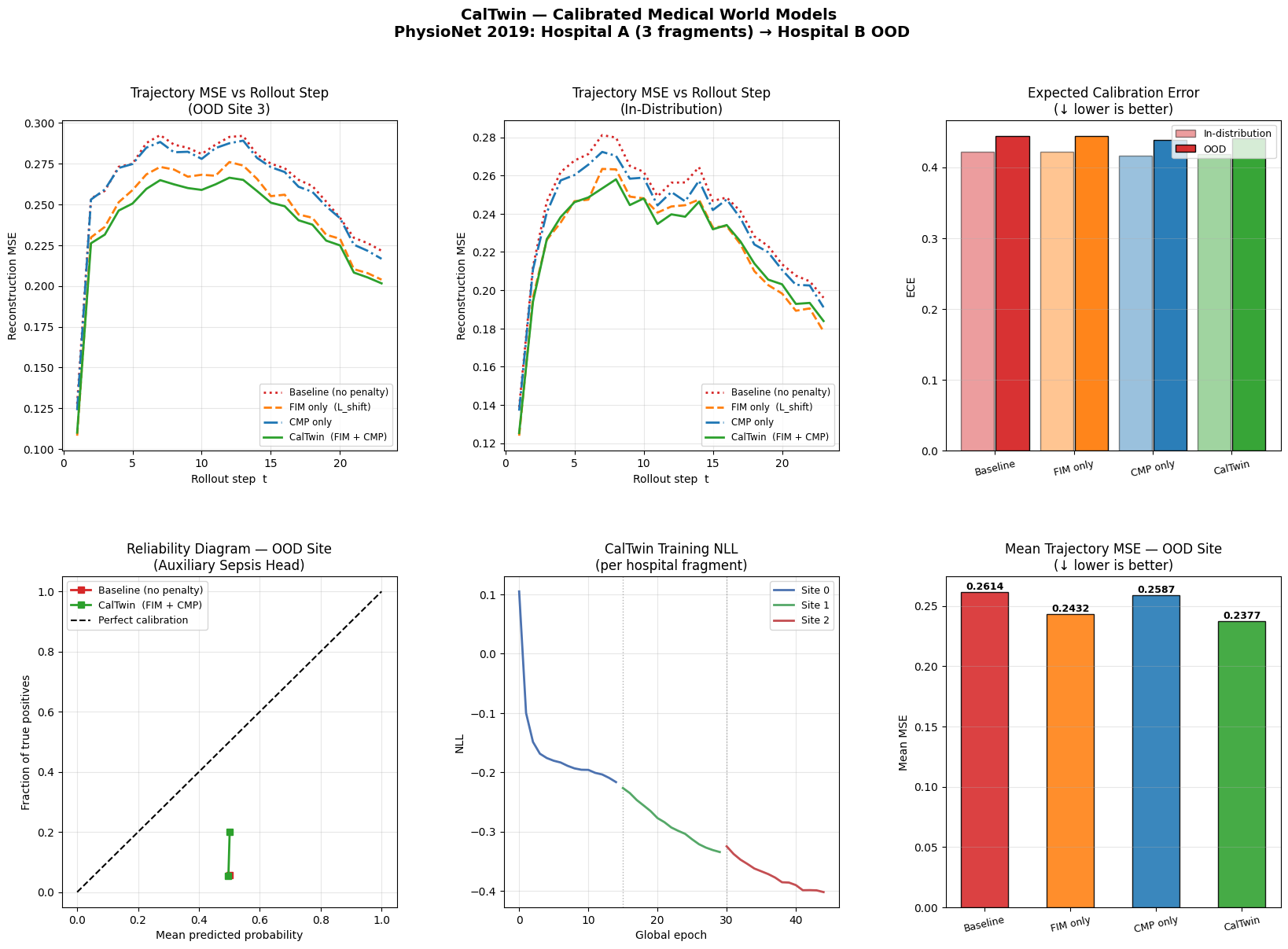}
    \caption{Empirical results on PhysioNet 2019 (single seed), all four
    methods of Table~\ref{tab:results}. Top: teacher-forced next-step MSE per
    rollout step (OOD, ID); ECE by method (ID vs.\ OOD). Bottom: reliability
    diagram for the auxiliary sepsis head, OOD, Baseline vs.\ CalTwin;
    CalTwin's training NLL across the three sequential fragments (dotted
    lines mark fragment boundaries); mean OOD MSE by method (= Table
    \ref{tab:results}'s MSE-OOD column, graphically).}
    \label{fig:results}
\end{figure}

Relative to baseline, CalTwin reduces OOD MSE by 9.1\% (FIM-only alone: 7.0\%)
and OOD ECE by 0.7\% (CMP-only alone: 1.3\%); the same ranking holds
in-distribution. FIM is thus the dominant contributor to the MSE improvement,
consistent with its anchoring role in Eq.~\ref{eq:lshift}, while CMP
contributes the (small) calibration gain.

Two limitations are directly visible in Table~\ref{tab:results}. First, once
the untrained-head artefact (Sec.~4.2) is accounted for, the CMP-trained
auxiliary head shows modest but genuine in-distribution discrimination
(AUROC $0.57$--$0.58$ vs.\ the $0.50$ untrained floor) that does
\emph{not} survive the move to the OOD hospital (AUROC $0.4754$--$0.4984$);
at this scale we cannot yet say whether this is a ceiling on the discrete-head
CMP adaptation or a data/capacity limitation. Second, because evaluation is
teacher-forced, these results characterise the one-step predictor under
fragment/hospital shift, not the autoregressive-rollout failure mode
motivated in Sec.~1; closing that gap is necessary before Sec.~1's
calibration claims are validated end-to-end.

\subsection{Second validation site: eICU-CRD Demo}

A single dataset is a real limitation for a MICCAI-workshop submission, so we
complete a second, independent validation here rather than deferring it to
camera-ready: Table~\ref{tab:eicu} reports actual results, run end-to-end on
a free-tier Google Colab GPU instance, from the protocol described
below  not a placeholder.

The eICU Collaborative Research Database Demo~\cite{pollard2018eicu} is a good fit:
unlike the full eICU-CRD or MIMIC-III/IV, it requires no PhysioNet
credentialing or CITI training (open-access, $\sim$50--130\,MB), and, unlike
PhysioNet 2019's two-hospital split, it is natively multi-site  2{,}500+ ICU
unit stays across 20 distinct hospitals (field \texttt{hospitalid})  so
fragments can be formed by \emph{real} hospital identity rather than an
enrolment-order proxy, directly extending the eICU/MIMIC-III shift
evidence~\cite{zhu2025fedweight} already motivating Sec.~1 and 4.1. Vitals in
\texttt{vitalPeriodic} (heart rate, $\mathrm{SpO}_2$, respiratory rate,
systolic/diastolic/mean BP) map directly onto the seven-channel input used
in Sec.~4.2, and in-hospital mortality supplies a binary outcome as a
drop-in replacement for the sepsis label used for CMP. The protocol mirrors
Sec.~4.1 exactly: group unit stays by \texttt{hospitalid}, select $K{=}3$
training hospitals as sequential fragments and one held-out hospital as the
OOD site, and reuse the Sec.~4.2 architecture and training loop unchanged.

\begin{table}[t]
\centering
\caption{\textbf{eICU-CRD Demo.} Performance on the eICU-CRD demo benchmark using the
same protocol and evaluation metrics as Table~\ref{tab:results} (hospital-fragment ID
split vs.\ held-out-hospital OOD split). Lower MSE and ECE are better, higher AUROC is
better (0.5 = chance). As in Table~\ref{tab:results} (Sec.~4.2), the auxiliary head is
trained only when $\lambda_2>0$, so AUROC-ID/OOD $=0.5000$ under Baseline/FIM-only
reflects an \emph{untrained} head. Unlike Table~\ref{tab:results}, CMP-only rather than
CalTwin attains the best value on four of six columns here (MSE-OOD, ECE-ID, ECE-OOD,
AUROC-OOD); we do not obscure this reversal (Sec.~5).}
\label{tab:eicu}
\resizebox{\textwidth}{!}{
\begin{tabular}{lcccccc}
\hline
Method & MSE-ID $\downarrow$ & MSE-OOD $\downarrow$ & ECE-ID $\downarrow$ & ECE-OOD $\downarrow$ & AUROC-ID $\uparrow$ & AUROC-OOD $\uparrow$ \\
\hline
Baseline (no penalty)
& 0.1480 & 0.1675 & 0.4867 & 0.4800 & 0.5000 & 0.5000 \\
FIM only ($\mathcal{L}_{\mathrm{shift}}$)
& 0.1478 & 0.1682 & 0.4867 & 0.4800 & 0.5000 & 0.5000 \\
CMP only
& 0.1433 & \textbf{0.1620} & \textbf{0.3811} & \textbf{0.3659} & 0.2635 & \textbf{0.8435} \\
\textbf{CalTwin (FIM+CMP)}
& \textbf{0.1428} & 0.1639 & 0.4342 & 0.4268 & 0.3446 & 0.8390 \\
\hline
\end{tabular}
}
\end{table}

Table~\ref{tab:eicu} both reinforces and complicates the PhysioNet picture.
It reinforces it in one respect: CalTwin again attains the best (lowest) ID
MSE, and the CMP-trained heads are again clearly distinguishable from the
untrained $0.5000$ floor. It complicates it in three respects that we state
precisely rather than average away. First, FIM-only does not improve OOD MSE
here  it is $0.4\%$ \emph{worse} than baseline ($0.1682$ vs.\ $0.1675$),
whereas on PhysioNet the identical penalty gave a $7.0\%$ reduction. Second,
CMP-trained AUROC-OOD is far \emph{above} chance on eICU ($0.8390$--$0.8435$)
rather than at or below it as on PhysioNet ($0.4754$--$0.4984$)  a large
effect in the opposite direction. Third, AUROC-ID for the trained heads is
\emph{below} chance on eICU ($0.2635$, $0.3446$), the reverse of PhysioNet's
above-chance ID discrimination. Consequently CMP-only, not CalTwin, attains
the best value on four of the six columns (MSE-OOD, ECE-ID, ECE-OOD,
AUROC-OOD); CalTwin's only outright win is MSE-ID. We return to what this
reversal does and does not license us to conclude in Sec.~5.

\section{Discussion}

Table~\ref{tab:results} supports three observations that qualify
Sections~1--3's claims on PhysioNet, stated explicitly rather than minimised.

\textbf{FIM drives the MSE gain; CMP's calibration gain is real but small,
and the two penalties interact.} FIM-only reduces OOD MSE by 7.0\% over
baseline, CalTwin extends this to 9.1\%, confirming the anchoring benefit of
Eq.~\ref{eq:lshift}. But CMP's OOD ECE improvement is \emph{smaller} under
CalTwin (0.7\%) than CMP-only (1.3\%)  the combined objective calibrates
worse than CMP alone here. A plausible mechanism: the FIM penalty constrains
the shared encoder's representation along directions useful for next-state
accuracy but not auxiliary-head calibration, limiting CMP's ability to
redistribute probability mass. Whether this is an artefact of fixed, untuned
$\lambda_1,\lambda_2$ or a structural property of the combined objective is
open without a hyperparameter search.

\textbf{CMP induces real but non-transferring discrimination in the
auxiliary head (on PhysioNet).} Since the head trains only via CMP (Sec.~4.2),
Baseline/FIM-only's exact $0.5000$ AUROC reflects an untrained head, not a
failure to learn. CMP-only and CalTwin reach AUROC $0.5731$/$0.5740$
in-distribution  modest but real, given a single linear readout trained 15
epochs/fragment on $\sim\!9\%$-prevalence data  but this collapses OOD to
$0.4984$/$0.4754$ (at or below chance). CalTwin's OOD AUROC is \emph{lower}
than CMP-only's, mirroring the ECE interaction above. What Table
\ref{tab:results}'s AUROC-ID column rules out is the reading that the head
never learns anything useful; the open question is specifically about
hospital-level transfer.

\textbf{Evaluation is teacher-forced, not closed-loop.} Table~\ref{tab:results}
characterises the one-step predictor under cross-fragment/-hospital shift,
with the true previous state fed at every step  not the compounding
autoregressive-rollout failure mode of Sec.~1, which requires closed-loop,
self-conditioned evaluation we leave to future work.

\textbf{The eICU-CRD result (Table~\ref{tab:eicu}) does not replicate
PhysioNet's ranking, and we treat that as informative rather than as noise to
explain away.} Three differences are large enough that a single-seed
fluctuation is an unsatisfying account of all of them simultaneously.
FIM-only helps OOD MSE on PhysioNet ($+7.0\%$) but not on eICU
($-0.4\%$, i.e.\ a slight loss), suggesting the FIM penalty's benefit is not
dataset-invariant  plausibly because eICU's fragmentation is by genuine
hospital identity (heterogeneous casemix, bed capacity, region) rather than
PhysioNet's within-system enrolment-order proxy, so the two experiments may
be probing different magnitudes, or even different kinds, of covariate
shift. CMP-trained AUROC-OOD is far above chance on eICU ($\approx\!0.84$)
versus at-or-below chance on PhysioNet ($\approx\!0.48$--$0.50$); a
plausible explanation is that 24-hour in-hospital mortality is an easier
signal to extract from ICU vitals than 24-hour sepsis onset, but we have not
tested this and state it as a hypothesis, not a finding. AUROC-ID for the
trained heads is below chance on eICU but above chance on PhysioNet, which
is harder to explain post hoc and is, at face value, more a symptom of
instability (a linear head trained on markedly fewer patients per fragment
than PhysioNet provides) than of either dataset's calibration properties. We
draw one methodological conclusion from this, not a scientific one:
single-dataset, single-seed evidence for a shift-robustness method is not
just incomplete but can be actively misleading about which component (FIM or
CMP) is doing the work, and we would not have known this without running the
second dataset.

Two further open questions remain. The empirical Fisher approximation
(Eq.~\ref{eq:empfim}) is known to diverge from the true Fisher under
misspecification~\cite{khan2025confidence}; whether this gap matters more for a
multi-step generative predictor than a single-step classifier is not
answerable from our results alone. And applying CMP to an auxiliary discrete
head, rather than the transition distribution directly, is a practical
compromise whose principled continuous-output alternative remains to be
derived and tested.

\section{Conclusion}

We have derived \textbf{CalTwin}, which applies Fisher-Information shift
regularisation and a Confidence Misalignment Penalty jointly to a medical
world model's latent transition predictor, and reported single-seed
feasibility studies on two real multi-hospital ICU datasets: PhysioNet 2019
(a documented instance of cross-hospital covariate
shift~\cite{zhu2025fedweight}) and the eICU-CRD Demo, fragmented by genuine
hospital identity. On PhysioNet, FIM reduces OOD next-step MSE by 7.0\% over
baseline and full CalTwin by 9.1\%; CMP's calibration gain is small (0.7\%
OOD ECE for CalTwin, 1.3\% for CMP alone), and the CMP-trained auxiliary head
shows modest in-distribution discrimination ($\text{AUROC}\approx 0.57$)
that does not survive the OOD hospital ($\approx 0.48$--$0.50$)  a transfer
failure, not an absence of learned signal. On eICU-CRD, the ranking does not
replicate: FIM-only does not improve OOD MSE, CMP-trained AUROC-OOD is far
above chance ($\approx\!0.84$) rather than at or below it, and CMP-only
rather than CalTwin wins on four of six metrics. We report both results
rather than the more favourable one, because the disagreement between them
is itself the most important finding of this feasibility study: it indicates
that FIM's and CMP's benefits, as instantiated here, are not yet
dataset-invariant, and that a method-contribution claim resting on either
dataset alone would have been unreliable. Both experiments characterise the
one-step predictor under teacher-forced evaluation, not the compounding
autoregressive-rollout failure mode motivating the method, and each rests on
a single seed.

Closing the gap to a clinically useful medical digital twin requires, at
minimum: running both datasets across multiple seeds to determine whether the
FIM/CMP reversal in Sec.~5 is systematic or an artefact of eICU's smaller
per-fragment sample size; tuning $\lambda_1,\lambda_2$ on held-out data rather
than using fixed values; evaluating under closed-loop self-conditioned
rollout; and moving from tabular ICU vitals to the imaging
modalities  cardiac, fetal ultrasound, surgical video  central to this
workshop. We offer CalTwin, both experiments, and this characterisation of
its transfer steps as a foundation for that work, and invite collaboration
with participants holding access to multi-site imaging data.

\bibliographystyle{splncs04}
\bibliography{ref}
\end{document}